%% file: main.tex
\documentclass[11pt]{article}
\usepackage{coling2020}
\usepackage{times}
\usepackage{url}
\usepackage{latexsym}

\usepackage{hyperref}
\usepackage{graphicx} 
\usepackage[utf8]{inputenc}
\usepackage{amsmath}
\usepackage{amsthm}
\usepackage{booktabs}
\usepackage{tikz}
\usepackage{pgfplots}
\usepackage{standalone}
\usepackage{subcaption}
\usepackage{paralist}
\usepackage{enumitem}
\usepackage{xspace}
\usepackage{listings}
\usepackage{amssymb}

\usepackage{microtype}

\usepackage{wrapfig}

\usepackage{wasysym}

\usetikzlibrary{calc}
\usetikzlibrary{positioning}

\usepackage[noend,noline]{algorithm2e}

\input{commands}

\colingfinalcopy 

\title{Generating Instructions at Different Levels of Abstraction}

\author{
  Arne Köhn\(^{*}\) \and Julia Wichlacz\(^{*}\) \and Álvaro Torralba\(^{**}\) \and \\
  \textbf{Daniel Höller\(^{*}\)} \and \textbf{Jörg Hoffmann\(^{*}\)} \and \textbf{Alexander Koller\(^{*}\)}
  \vspace{1ex} \\
  \begin{minipage}[t]{0.55\linewidth}
    \centering
    \(^{*}\)Saarland Informatics Campus\\
    Saarland University\\
    \{\href{mailto:koehn@coli.uni-saarland.de}{koehn},
    \href{mailto:koller@coli.uni-saarland.de}{koller}\}@coli.uni-saarland.de\\
    \{\href{mailto:wichlacz@cs.uni-saarland.de}{wichlacz},
    \href{mailto:hoeller@cs.uni-saarland.de}{hoeller},
    \href{mailto:hoffmann@cs.uni-saarland.de}{hoffmann}\}@cs.uni-saarland.de
  \end{minipage}
  \hspace{-4em}
  \begin{minipage}[t]{0.45\linewidth}
    \centering
    \(^{**}\)Department of Computer Science\\
    Aalborg University\\
    \href{mailto:alto@cs.aau.dk}{alto@cs.aau.dk}
  \end{minipage}
}
\date{}

\begin{document}
\maketitle

\begin{abstract}
  When generating technical instructions, it is often convenient to
  describe complex objects in the world at different levels of
  abstraction. A novice user might need an object explained piece by
  piece, while for an expert, talking about the complex object (\eg{}
  a wall or railing) directly may be more succinct and efficient. We
  show how to generate building instructions at different levels of
  abstraction in Minecraft. We introduce the use of hierarchical
  planning to this end, a method from AI planning which can capture
  the structure of complex objects neatly. A crowdsourcing evaluation
  shows that the choice of abstraction level matters to users, and
  that an abstraction strategy which balances low-level and high-level
  object descriptions compares favorably to ones which don't.

\end{abstract}

\input{introduction}

\input relwork

\input{approach}

\input{construction}

\input{abstraction-level}

\input{experiments}

\input{conclusion}

\paragraph{Acknowledgements} Funded by the Deutsche
Forschungsgemeinschaft (DFG, German Research Foundation) –
Project-ID 232722074 – SFB 1102.

\bibliographystyle{coling}
\bibliography{coling2020}

\appendix

\input appendix

\end{document}

%% file: commands.tex
\definecolor{CommentColor}{rgb}{0.90,0.16,0} 

\ifdefined\nocflagdefined

\newcommand{\joerg}[1]{}
\newcommand{\julia}[1]{}
\newcommand{\daniel}[1]{}
\newcommand{\alvaro}[1]{}
\newcommand{\alex}[1]{}
\newcommand{\arne}[1]{}

\else

\newcommand{\joerg}[1]{\textbf{\color{CommentColor} /* (Joerg) #1  (Joerg) */}}
\newcommand{\julia}[1]{\textbf{\color{CommentColor} /* (Julia) #1  (Julia) */}}
\newcommand{\daniel}[1]{\textbf{\color{CommentColor} /* (Daniel) #1  (Daniel) */}}
\newcommand{\alvaro}[1]{\textbf{\color{CommentColor} /* (Alvaro) #1  (Alvaro) */}}
\newcommand{\alex}[1]{\textbf{\color{CommentColor} /* (Alex) #1  (Alex) */}}
\newcommand{\arne}[1]{\textbf{\color{CommentColor} /* (Arne) #1  (Arne) */}}

\fi

\newcommand{\ie}{i.\,e.}
\newcommand{\eg}{e.\,g.}

\newcommand{\factName}[1]{\textit{#1}}
\newcommand{\taskName}[1]{\texttt{#1}}
\newcommand{\abstractTaskName}[1]{\textsc{#1}}

\newcommand{\boldAndIt}[1]{\textbf{\textit{#1}}}
\newcommand{\HObjects}{{\ensuremath{\cal O}}} 

\newcommand{\FSet}{\ensuremath{F}} 
\newcommand{\pTaskNames}{\ensuremath{A}} 
\newcommand{\cTaskNames}{\ensuremath{C}} 
\newcommand{\taskNetwork}{\ensuremath{\mathit{tn}}} 
\newcommand{\decmethods}{\ensuremath{M}} 
\newcommand{\sinit}{\ensuremath{s_0}} 

\newcommand{\precond}{\ensuremath{\mathit{prec}}} 
\newcommand{\add}{\ensuremath{\mathit{add}}} 
\newcommand{\del}{\ensuremath{\mathit{del}}} 

\newcommand{\FSetInstruction}{\ensuremath{F}\xspace} 
\newcommand{\pTaskNamesInstruction}{\ensuremath{A}\xspace} 
\newcommand{\cTaskNamesInstruction}{\ensuremath{C}\xspace} 
\newcommand{\decmethodsInstruction}{\ensuremath{M}\xspace} 
\newcommand{\tniInstruction}{\ensuremath{\taskNetwork_I}\xspace} 

\newcommand{\KnowledeLevelLow}{Low-level\xspace} 
\newcommand{\KnowledeLevelMiddle}{Teaching\xspace} 
\newcommand{\KnowledeLevelHigh}{High-level\xspace} 

%% file: introduction.tex
\section{Introduction}
\label{sec:introduction}

Technical instructions in complex environments can
often be stated at different levels of
abstraction. For instance, a natural language generation (NLG) system
for tech support might instruct a human instruction follower (IF) to
either plug ``the broadband cable into the broadband filter'' or ``the
thin white cable with grey ends into the small white box''
\cite{janarthanam10:_learn_adapt_unknow_users}. Depending on how much
the IF knows about the domain, the first, high-level instruction may
be difficult to understand, or the second, detailed instruction may be
imprecise and annoyingly verbose. An effective instruction generation
system will thus adapt the level of abstraction to the user.

\begin{figure}
  \centering
$\vcenter{\hbox{\includegraphics[width=0.48\textwidth]{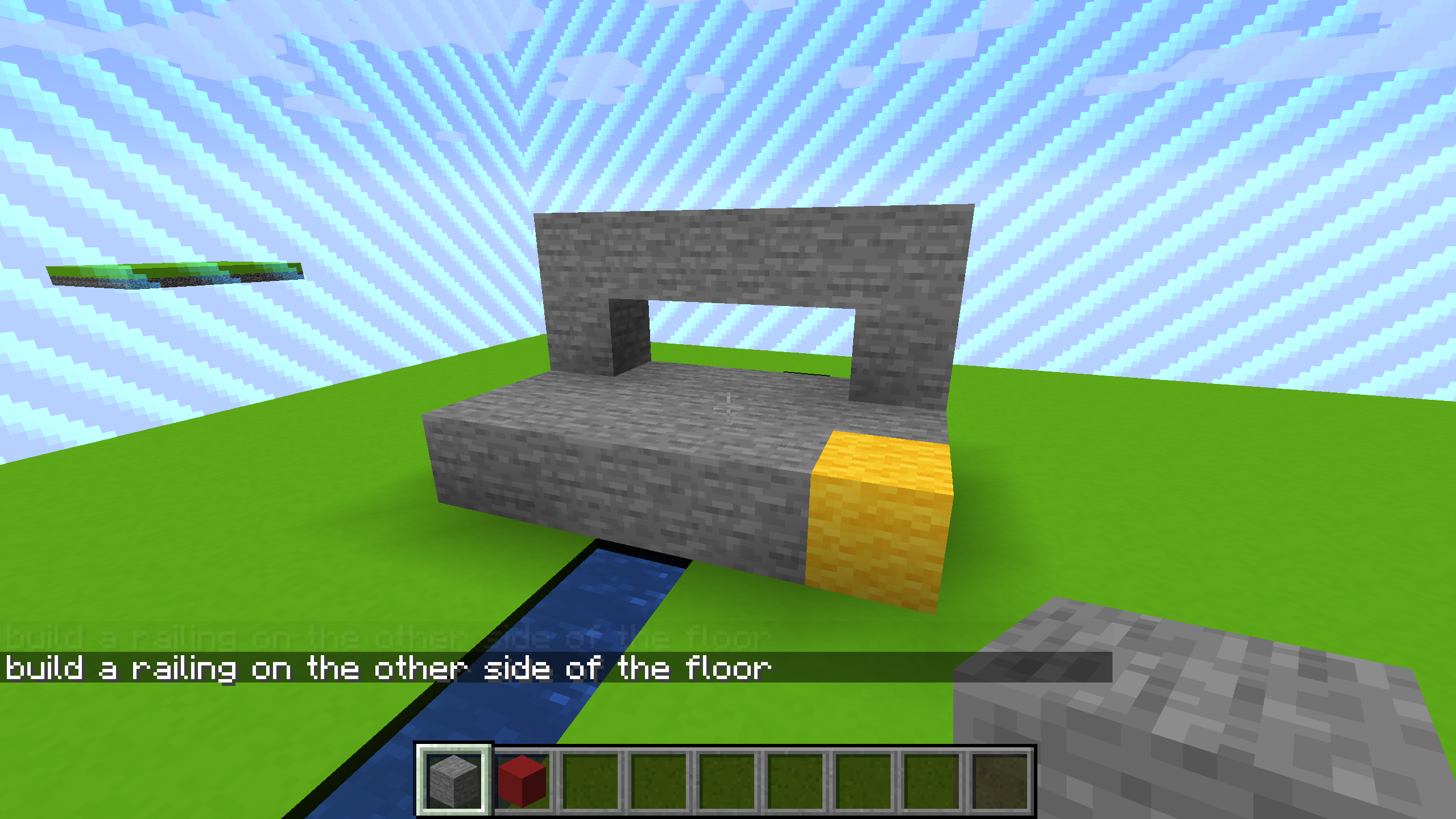}}}$\hfill
\begin{minipage}{0.5\textwidth}
  \small
    Instructing the railing (here already built):
  \begin{compactdesc}
  \item[(a)] ``Put a block on top of the blue block. Put a block on top of the previous block. [\ldots]''
  \item[(b)] ``Now I will teach you how to build a railing. Put a block  on top of the blue block. [\ldots]''
  \item[(c)] ``Build a railing from the top of the blue block to the back right corner of the floor.''
  \end{compactdesc}
  \vspace{1em}
  Instructing a second railing to be built:
  \begin{compactdesc}
  \item[(d)] ``Put a block on top of the yellow block. Put a block on top of the previous block. [\ldots]''
  \item[(e)] ``Build a railing on the other side of the floor.''
  \end{compactdesc}
\end{minipage}%
\caption{Different possibilities to instruct the IF to build a part of a bridge.}
  \label{fig:railing}
\end{figure}

In this paper, we investigate the generation of instructions at
different levels of abstraction in the context of the computer game
``Minecraft''.
Minecraft offers a virtual 3D environment in which the player can mine
materials, craft items, and construct complex objects, such as
buildings and machines. The user constructs these objects bit by bit
from atomic blocks, and it is always possible to generate
natural-language instructions which describe the placement of each
individual block. However, it can be more effective to generate more
high-level instructions. In the bridge-building example shown in
Fig.~\ref{fig:railing},
it is probably better to simply say ``build a railing on the other
side'' instead of explaining where to place the seven individual
blocks -- provided the IF knows what a railing looks like. 
Minecraft is the best-selling video game of all time (200 million
users), which means that there is a large pool of potential
experimental subjects for evaluating NLG systems. Minecraft has been
used previously as a platform for experimentation in AI and in
particular for NLG
\cite{narayan-chen-etal-2019-collaborative,kohn-koller-2019-talking}.

We present an instruction giving (IG) system which guides the user in
constructing complex objects in Minecraft, such as houses and
bridges. The system consists of two parts: a hierarchical planning
system based on \emph{Hierarchical Task Networks (HTN)}
\cite{GhallabBook,bercher:etal:ijcai19}, which computes a structured
instruction plan for how to to explain the construction;
and a chart-based generation system which inputs individual plan steps
and generates the actual instruction sentences
\cite{kohn-koller-2019-talking}.

Planning systems generate plans based on expressive declarative models,
making this approach easily
applicable to a wide range of domains, and giving it the power to deal
with large degrees of freedom in instruction generation. 
In particular, we leverage the hierarchical planning system to obtain
three different strategies for describing complex objects as
illustrated in Fig.~\ref{fig:railing}: \emph{low-level}, always
instructing block-by-block (sentences (a) and (d)); \emph{high-level},
always instructing to build the next complex subobject (sentences (c)
and (e)); and a \emph{teaching} strategy which first explains how to
construct a complex subobject, and then uses high-level descriptions
for that object in subsequent instructions (sentences (b) and (e)). We
realize these strategies through designing the planner's action-cost
function.
The strategies constitute a first step towards a fully generic IG
system which adapts its description of complex objects to the user's
knowledge. The planner's cost function could also be used to
incorporate additional criteria, \eg{} that the generated sentences
should be easy to understand in context.

We evaluate our IG system by crowdsourcing, using the open-source
MC-Saar-Instruct platform for IG experiments in the Minecraft domain
\cite{kohn20:_mc_saar_instr}. Each user receives instructions for
building either a house or a bridge using one of the three abstraction
strategies sketched above. The results show significant differences in
completion times and user satisfaction across the three strategies. In
the bridge scenario, the teaching strategy outperforms the high-level
strategy in completion time and the low-level strategy in user
satisfaction. In the house scenario, the high-level
strategy outperforms the others in completion time for the walls,
illustrating the importance of modeling user knowledge when
choosing the level of abstraction.

\paragraph{Plan of the paper.} After reviewing some related work
(Section~\ref{sec:relwork}), we overview the architecture of our IG
system (Section~\ref{sec:approach}). We briefly introduce HTN
planning, and how we use it to define hierarchical Minecraft
construction planning models (Section~\ref{sec:construction}). We then
explain how to get from \emph{construction} planning models to
\emph{instruction} planning models, and how to generate instruction
plans at different levels of abstraction
(Section~\ref{sec:abstraction-level}). Section~\ref{sec:experiments}
describes our evaluation; Section~\ref{sec:conclusion}
concludes.\footnote{Software and data are available at
  \url{https://minecraft-saar.github.io/coling2020}}

%% file: relwork.tex
\section{Related Work}
\label{sec:relwork}

Generating natural-language instructions grounded in the mechanics of
a real or virtual world is a well-established type of NLG task. For
instance, the GIVE Challenge \cite{koller-etal-2010-give} required NLG
systems to guide human users through a maze while referring to
locations and objects. We follow the GIVE Challenge in situating the
task in a virtual environment and using crowdsourced task-based
evaluation, but the Minecraft world is much more complex than the GIVE
world, and contains complex objects.

\newcite{rookhuiszen09:_two_give} describe an IG system for the GIVE
Challenge which dynamically adapts the level of detail of navigation
instructions. They use a simple heuristic to switch between
abstraction levels.  \newcite{janarthanam10:_learn_adapt_unknow_users}
generate referring expressions at different levels of
abstraction in an electronics repair scenario, and adapt to novice
vs.\ expert users. They assume a finite set of possible descriptions
for each object, and do not exploit the internal structure of complex
objects.

The use of AI planning for elaborating and organizing the things that
need to be said in an NLG system (\ie{} \emph{discourse planning})
dates back into the early days of NLG
\cite{appelt85:_plann_englis_senten,hovy88:_gener}. \newcite{garoufi-koller-2014-effect-ref-express-sit-cont}
use non-hierarchical planning to compute communicative plans. They use
planning operators which encode communicative actions, and allow them
to have effects both on the communicative state and on the state of
the world; we disentangle these effects in our hierarchical planning
model. \newcite{BehnkeBKSMHDDMM20} use a hierarchical planner to
generate technical instructions at different levels of abstraction,
but their system can only utter sentences which were stored with the
planning operators as canned text. Their evaluation does not show that
users prefer their system over a baseline, illustrating the difficulty
of generating instructions at the right abstraction level.

The Minecraft domain has been used extensively for various tasks in AI
\cite{aluru15,parashar17}, including planning
\cite{roberts:etal:icaps-ws-17}
and natural language understanding
\cite{gray2019craftassist}. Regarding NLG specifically,
\newcite{narayan-chen-etal-2019-collaborative} trained a neural model
to generate building instructions in Minecraft; in the absence of
symbolic domain knowledge, their model struggles to generate correct
instructions. \newcite{kohn-koller-2019-talking} show how individual
instructions can be generated in Minecraft. Their focus is on
generating indefinite referring expressions to objects which do not
exist yet because the user is supposed to build them. Here we do not
address how to generate the individual utterances, but how to
determine the semantic content of these utterances.

%% file: approach.tex
\section{NLG System Architecture}
\label{sec:approach}

Our overall IG system consists of two separate modules: an
\emph{instruction planner} and a \emph{sentence generator}. The role
of the instruction planner is to compute an \emph{instruction plan},
\ie{} a sequence of \emph{instruction actions}. An instruction action
is an abstract semantic representation of a sentence; the sentence
generator then translates it into a natural-language utterance, such
as ``place a block on top of the yellow block'', ``build a floor from
the black block to the yellow block'', or ``now I will teach you how
to build a railing''.

The technical focus of this paper is on the instruction planner.
Given a state of the Minecraft world and 
a
specification for the complex object the IF is supposed to build (\eg{} a
bridge or a house), the instruction planner will compute a sequence of
instruction actions as explained above.
One technical contribution of this paper is that the instruction
actions can be at different levels of abstraction; for example, the
instruction plan can simply say ``build a railing on the other side''
in the situation of Fig.~\ref{fig:railing}, or it can explain how to
build the railing block by block.
To ensure the correctness of the instruction plan, \ie{} to ensure
that the instructions, followed correctly, actually do result in the
intended complex object, the instruction planner performs
\emph{construction} planning as part of its planning process:
it internally refines its high-level actions into block-by-block plans
and checks that these work.

The sentence generator takes instruction actions as input and produces
natural-language utterances. We use the chart generation system of
\newcite{kohn-koller-2019-talking} to generate sentences. This system
generates sentences while simultaneously generating definite and
indefinite referring expressions (REs); definite REs are used to refer
to objects which already exist in the Minecraft environment, and
indefinites are generated to refer to objects which do not exist yet
because the IF is supposed to build them. Thus for instance, the
sentence generator might translate the instruction action
\taskName{ins-railing}(0,1,4,5,south) to ``build a railing from the
top of the blue block to the top of the red block'' or ``build a
railing on the other side of the floor'', depending on the state of the dialogue
and the Minecraft world.

%% file: construction.tex
\section{Hierarchical Construction Planning}
\label{sec:construction}

As indicated, we distinguish between \emph{construction} planning,
which determines how a complex object can be built block-by-block; and
\emph{instruction} planning, which determines how, and in particular
at which level of abstraction, the building of a complex object can be
explained to the IF. Instruction planning encompasses construction
planning as a sub-problem. We employ \emph{hierarchical} planning for
both tasks, with hierarchies of complex subobjects and the associated
building activities. These are not required for construction planning
per se (which can always proceed on a block-by-block basis).  But they
speed up the construction planning
process, and they are key to instruction planning as proposed here.
Previous work on planning in Minecraft \cite{roberts:etal:icaps-ws-17}
has considered construction planning only, and has not considered
hierarchical planning. 
We now introduce our construction planning models, which we will
extend to instruction planning models in
Section~\ref{sec:abstraction-level}.

\subsection{Background: HTN Planning}
\label{sec:construction:htn}

Hierarchical Task Network (HTN) planning (see
\newcite{bercher:etal:ijcai19} for a recent overview) comes with
different levels of abstraction regarding the things that must be
done, the \emph{tasks}. \emph{Primitive tasks} (also \emph{actions})
can directly be
executed in the environment. They come with conditions that need to
hold to make them applicable, and their application changes the
environment.
\emph{Abstract tasks} describe behavior at a higher level of
abstraction.  They are not applicable directly, but must
instead be divided into other tasks by using \emph{decomposition
  methods}. The new tasks may, again, be abstract or
primitive. Methods are similar to derivation rules in a formal
grammar where the left-hand side the abstract task and the
right-hand side is its decomposition into other tasks/actions. Here
we use \emph{totally-ordered} HTN planning, a common subclass where
the right-hand side is restricted to be a task sequence.
Planners are given the overall tasks to accomplish, \eg\ \emph{build a
  bridge}. These are decomposed until only actions are left, which
must be applicable in the initial state of the system.

We now give a definition using the basic formalism introduced by
\cite{BehnkeHB18}.  An HTN planning problem $\mathcal{P}$ is a tuple
$(\FSet,\cTaskNames,\pTaskNames,\decmethods,\taskNetwork_I,\sinit)$:
\begin{compactitem}
 \item $\FSet$ is a set of propositional state features used to
   describe the environment. A \emph{state} $s$ is a truth assignment
   to these features, usually represented by the set of features
   true in the state.
 \item $\cTaskNames$ and $\pTaskNames$ are sets of abstract (also
   \underline{c}ompound) tasks and primitive tasks (also
   \underline{a}ctions).
 \item $\decmethods \subseteq \cTaskNames \times (\cTaskNames \cup
   \pTaskNames)^*$ is the set of decomposition methods (where $^*$
   is the Kleene operator).
 \item $\taskNetwork_I = (\cTaskNames \cup \pTaskNames)^*$ is the
   initial task network, $\sinit \in 2^{\FSet}$ is the initial state
   of the environment.
\end{compactitem}
Furthermore $\mathcal{P}$ is associated with functions $\precond$,
$\add$, and $\del$ that map each action to its preconditions,
add-effects, and delete-effects. \precond{} is a logical formula over
$\FSet$ such that an action $a$ is applicable if $\precond$ is
satisfied in current state $s$.
When $a$
is applicable in $s$, the state resulting from its application is
defined as $\left(s\setminus \del(a) \right) \cup \add(a)$ where
$\add, \del \subseteq F$.
The sets of all possible states and actions (implicitly) define a
state transition system describing how the environment can change.

Plans in HTN are defined through \emph{task networks}, which 
are sequences in $(\cTaskNames \cup \pTaskNames)^*$.
If $\taskNetwork = \omega c\, \omega''$ with $c \in \cTaskNames$ is a
task network and $m = (c, \omega') \in \decmethods$ is a method, then
$\taskNetwork$ can be decomposed with $m$ into $\taskNetwork' =
\omega\, \omega' \omega''$.
A plan $\taskNetwork_S$ is a sequence in $\pTaskNames^*$ that can be
obtained by iteratively decomposing the initial task network, and that
is applicable in the initial state.
Plan quality is measured in terms of a cost function, $\mathit{cost} :
2^F \times A \rightarrow \mathbb{R}^+_0$, where the task of a planner
is to minimize the summed up cost of the resulting plan.

Deciding plan existence in the framework we use here is, in general,
EXPTIME-complete \cite{ErolHN94,AlfordBA15}. A range of solvers is available that
tackles this complexity through search
\cite{Nau:etal:jair03,Bercher:etal:ijcai17,holler:etal:jair20} or
compilation into simpler frameworks
\cite{alford:etal:ijcai-09,behnke:etal:aaai-19}. Some solvers provide
guarantees on the solution costs ~\cite{behnke:etal:ijcai-19}, or
enable anytime behavior continuing search to find better
plans~\cite{wichlacz:etal:socs20}.

\subsection{HTN Construction Planning Models for Minecraft}
\label{sec:construction:htnminecreaft}

As a compact explanation of our HTN construction planning models,
consider the part of Fig.~\ref{fig:complete_model} that is
highlighted in boldface, which shows a construction model for
building a bridge. The propositional state features $\FSet$ take the
form \factName{block(x,y,z)}, encoding whether or not there is a block
at those coordinates. The construction of any complex object in
Minecraft can ultimately be decomposed into
\taskName{put-block(x,y,z)} actions, which correspond to the primitive
tasks in the HTN model.\footnote{In general, every \taskName{put-block(x,y,z)} action
  would need to be guarded with a precondition that a block can be placed at (x,y,z), \ie{} that it is
  not placed in thin air). In our case, such preconditions are not necessary since the task hierarchy
  ensures this.}

While all our Minecraft construction models share the same state
features and primitive actions, they differ on the complex objects
that can be built, {\HObjects}. In the case of a bridge, these are the
floor, the railings, and rows of blocks, placed at different positions
and in different orientations.  Each complex object $X$ has an
associated abstract task \abstractTaskName{build-X}. For example,
the task \abstractTaskName{build-bridge(0,0,0,5,3,north)}
corresponds to building a bridge of size 5x3 facing north and starting
at position 0,0,0. A construction model specifies one or more
decomposition methods for each  abstract task, corresponding
to different ways to build the
object.  This is illustrated in Fig.~\ref{fig:complete_model} by the two
decomposition methods for \abstractTaskName{build-railing}. The HTN
planning system, run on such a model, will choose which option is more
suitable for the task at hand to minimize the overall action cost. In
Section~\ref{sec:abstraction-level:cost}, we will specify cost
functions suited to optimize instructions for an IF.

\begin{figure}[t]
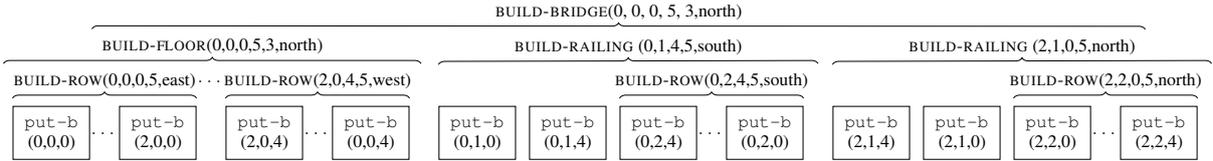

  \includestandalone[width=\textwidth]{tikz/bridge}
  \caption {Example of an HTN plan for building a bridge of width 3, and length 5 at
    position (0, 0, 0).}
  \label{fig:instruction_plan_diagram:low-level}
\end{figure}

Fig.~\ref{fig:instruction_plan_diagram:low-level} shows a
plan. \abstractTaskName{build-bridge}(0,0,0,5,3,north) is decomposed
into a floor and two railings. Other decompositions are possible,
\eg{}, constructing the railings in different order or
direction. The tasks are further decomposed until a
valid sequence of \taskName{put-block} actions is reached.

Minecraft construction models -- in general, and in particular the
models we devise here -- are challenging for HTN planning systems due
to the large number of objects required to characterize a 3D world. In
our experiments, we use recent algorithms based on Monte-Carlo Tree
Search, which do not require a grounding pre-process
and are thus able to scale to
comparatively large Minecraft worlds while optimizing plan
cost~\cite{wichlacz:etal:socs20}.

%% file: tikz/bridge.tex
\tikzstyle{constructionstep}=[draw,  rectangle, text width=0.8cm, align=center, font=\scriptsize]

\begin{tikzpicture} [node distance = 1.25cm] 
  \node [constructionstep] (C1) at (-0.5,0.7) {\taskName{put-b}\\ (0,0,0)};
  \node [constructionstep] (C2) [right = 0.4cm of C1] {\taskName{put-b}\\ (2,0,0)};
  \node []  at ($(C1)!0.5!(C2)$) {\tiny $\dots$};
  \draw[decoration={brace,raise=3pt},decorate] (C1.north west) -- node[above=3pt,font=\scriptsize] (R1) {\abstractTaskName{build-row}(0,0,0,5,east)} (C2.north east);

  \node [constructionstep] (C3)  [right = 0.4cm of  C2]  {\taskName{put-b}\\ (2,0,4)};
  \node [constructionstep] (C4) [right = 0.4cm of  C3] {\taskName{put-b}\\ (0,0,4)};
  \node []  at ($(C3)!0.5!(C4)$) {\tiny $\dots$};
  \draw[decoration={brace,raise=3pt},decorate] (C3.north west) -- node[above=3pt,font=\scriptsize] (R2) {\abstractTaskName{build-row}(2,0,4,5,west)} (C4.north east);
  \draw[decoration={brace,raise=0pt},decorate] (R1.north west) -- node[above=0pt,font=\scriptsize] (Floor) {\abstractTaskName{build-floor}(0,0,0,5,3,north)} (R2.north east);
  \node []  at ($(R1)!0.5!(R2)$) {\tiny $\dots$};

  \node [constructionstep] (C5)  [right = 0.4cm of  C4]  {\taskName{put-b}\\ (0,1,0)};
  \node [constructionstep] (C52)  [right of = C5]  {\taskName{put-b}\\ (0,1,4)};
  \node [constructionstep] (C6)  [right of =  C52]  {\taskName{put-b}\\ (0,2,4)};
  \node [constructionstep] (C7) [right = 0.4cm of C6] {\taskName{put-b}\\ (0,2,0)};
  \node []  at ($(C6)!0.5!(C7)$) {\tiny $\dots$};
  \draw[decoration={brace,raise=3pt},decorate] (C6.north west) -- node[above=3pt,font=\scriptsize] (R3) {\abstractTaskName{build-row}(0,2,4,5,south)} (C7.north east);
  \draw[decoration={brace ,raise=0pt},decorate] (C5.north west) + (0,0.575) -- node[above=0pt,font=\scriptsize] (Rail1) {\abstractTaskName{build-railing} (0,1,4,5,south)} (R3.north east);

  \node [constructionstep] (C8)  [right = 0.4cm of  C7]  {\taskName{put-b}\\ (2,1,4)};
  \node [constructionstep] (C82)  [right of = C8]  {\taskName{put-b}\\ (2,1,0)};
  \node [constructionstep] (C9)  [right of =  C82]  {\taskName{put-b}\\ (2,2,0)};
  \node [constructionstep] (C10) [right = 0.4cm of  C9] {\taskName{put-b}\\ (2,2,4)};
  \node []  at ($(C9)!0.5!(C10)$) {\tiny $\dots$};
  \draw[decoration={brace,raise=3pt},decorate] (C9.north west) -- node[above=3pt,font=\scriptsize] (R4) {\abstractTaskName{build-row}(2,2,0,5,north)} (C10.north east);

  \draw[decoration={brace ,raise=0pt},decorate] (C8.north west) + (0,0.575) -- node[above=0pt,font=\scriptsize] (Rail2) {\abstractTaskName{build-railing} (2,1,0,5,north)} (R4.north east);

  \draw[decoration={brace,raise=0pt},decorate] (Floor.north west) -- node[above=0pt, font=\scriptsize] (H1) {\abstractTaskName{build-bridge}(0, 0, 0, 5, 3,north)} (Rail2.north east);

\end{tikzpicture}

%% file: abstraction-level.tex
\section{From Construction Planning to Instruction Planning}
\label{sec:abstraction-level}

Having discussed how to compute a \emph{construction} plan with an HTN
planner, we will now explain how to compute \emph{instruction}
plans. The actions in an instruction plan represent communicative
actions, in which the IG system sends a sentence to the IF. These
actions can describe the intended construction steps
at different levels of abstraction. We capture this in
the HTN model by defining a task for instructing the IF to build each
complex subobject, say a railing. The planner can then choose to
either achieve this task with a single primitive instruction action
(which might send ``build a railing'' to the IF), or to further
decompose it into smaller tasks, which will instruct the IF to build
two blocks and connect them with a row. 

One key challenge is that while the instruction actions represent
natural-language instructions, they still need to be grounded in
activities in the Minecraft world: we must guarantee that,
if the IF follows the instructions correctly, then
the correct complex object results. To this end, we reason
about the construction and its explanation simultaneously.
For every instruction action, the plan also contains the
corresponding block-by-block construction actions, thus validating
the instruction plan.

\input htn-model-complete

\subsection{HTN Instruction Planning Model}
\label{sec:abstraction-level:htn}

We extend the construction model of
Section~\ref{sec:construction:htnminecreaft} to an instruction model by adding
the non-boldface parts of Fig.~\ref{fig:complete_model}.
Since we plan for instruction giving, we have to consider the IF and
their knowledge. Some complex objects might be known to the IF (\eg{}
what a row of blocks is), others might not be. For example, if we
instruct the IF to build the first railing for the bridge depicted in
Fig.~\ref{fig:railing}, the IF most likely does not know the exact
shape this railing should have. Thus, we introduce a propositional
state feature \factName{knows-T} (see \FSetInstruction in
Fig.~\ref{fig:complete_model}) for each kind of complex object \(T\),
representing whether or not the IF knows how to build such objects.
We also incorporate information about the block the IF was instructed
to place last. This is used by the cost function (see
Section~\ref{sec:abstraction-level:cost}) to take into account that
referring expressions are easier to generate, and to understand, for
positions adjacent to the last placed block.

The instruction model also has new primitive actions $\pTaskNames$,
whose names start with \taskName{ins-}; executing such an action
corresponds to generating a sentence and sending it to the IF. First,
for every instance \(X\) of a complex subobject, and for every
possible block \(X\), there is an action \taskName{ins-X} that
represents asking the IF to build \(X\). For complex subobjects, this
action has the precondition that the IF knows the corresponding
high-level concept. Second, the actions \taskName{ins-teach-start-X}
and \taskName{ins-teach-end-X} correspond to utterances like ``I will
now teach you how to build a railing''. These teaching actions have no
preconditions and add \taskName{knows-X} to the state.

We furthermore add compound tasks \abstractTaskName{ins-build-X} into
$\cTaskNames$, which coordinate the instruction for object \(X\),
through different decomposition methods in $\decmethods$. These
methods play a crucial role in our approach, as they allow the
modeller to encode different possible instruction variants. Here, we
explore this modelling power by allowing each
\abstractTaskName{ins-build-X} task to decompose in three different
ways, corresponding to different levels of abstraction in the
explanation of a complex object:

\begin{compactitem}
\item Decomposition L chooses to explain a complex object in terms of
  its parts. For instance, it decomposes an ``instruct to build
  railing'' task into two ``instruct to place block'' actions and an
  ``instruct to build row'' task. Always choosing Decomposition L
  results in low-level block-by-block instructions. The plan will
  alternate \taskName{ins-block} actions (instructing the user to
  ``place a block'') with \taskName{put-block} actions (ensuring
  correctness through construction planning).
\item Decomposition H generates a single instruction action for the
  object at the current level of abstraction. For the railing, it uses
  the primitive instruction action \taskName{ins-railing} (``build a
  railing''). This action has a precondition \factName{knows-railing},
  \ie{} this decomposition can only be chosen if the IF already
  understands the concept of a railing (either by initial expertise or
  by previous teaching through decomposition T, see next). The
  instruction is followed by the construction-planning task
  \abstractTaskName{build-railing} ensuring correctness.
\item Decomposition T defers instruction to the lower level (like L),
  but also adds instruction actions \taskName{ins-teach-start-X} and
  \taskName{ins-teach-end-X}. These generate utterances like ``I will
  teach you how to build a railing'' and have the effect that the IF
  now understands the corresponding high-level concept (\eg{} making
  the precondition \factName{knows-railing} of \taskName{ins-railing}
  true) so that, later on, decomposition H can be used.
\end{compactitem}

\input{instruction-plan-example}

\subsection{Designing the Cost Function}
\label{sec:abstraction-level:cost}

Given an HTN instruction-planning model as just defined, the HTN
planner automatically decides which decomposition methods (in
particular: L/H/T) are used at which points of the instruction, in a
manner that minimizes plan cost. Instruction quality can thus be
optimized by suitably defining the cost function. We demonstrate this
by showing how the cost function, in combination with the initial
environment state $\sinit$, can be used to realize different
instruction-giving \emph{strategies}. In other words, we show how the
HTN planner can choose the appropriate level of abstraction.
We have realized the following three strategies:

\begin{compactdesc}
\item[\KnowledeLevelLow{}] This strategy always explains how to build
  complex objects block by block. The initial
  state does not contain any \factName{knows-X} features, and the
  cost function assigns a high cost to instruction actions for
  complex objects (\taskName{ins-X} and \taskName{ins-teach-X}). Thus,
  minimal-cost plans always use the L decomposition.
  (cmp.\ Fig.~\ref{fig:railing} a and d)

\item[\KnowledeLevelHigh{}] This strategy always explains how to build
  complex objects with a single abstract instruction.
  All \factName{knows-X} state features are true in the initial state, \ie{},
  the IF is assumed to know how to build all complex objects. The
  \taskName{ins-X} actions are assigned low costs, relative to that of
  \taskName{ins-block} actions.  Therefore, minimal-cost plans always use the
  H decomposition.  (cmp.\ Fig.~\ref{fig:railing} c and e)

\item[\KnowledeLevelMiddle{}] This strategy strikes a balance between
  the first two. Its initial state assumes that all \factName{knows-X}
  state features are false; hence, like \KnowledeLevelLow{}, it explains each
  complex object in simpler terms when it is first built. However, it
  encourages the use of decomposition T to do this, so that for later
  instances of the same object decomposition H can be used. This is
  achieved by assigning a low cost to \taskName{ins-X} and
  \taskName{ins-teach-X} relative to the cost of \taskName{ins-block}.
  (cmp.\ Fig.~\ref{fig:railing} b and e)
\end{compactdesc}

These strategies could, of course, be implemented without the use of a
general HTN planning system. However, our generic implementation
readily handles deeper user models if available, and it facilitates
the flexible combination with other criteria. In particular, the cost
of \taskName{ins-block} actions can be used to model what the sentence
generator can or cannot express easily. In general, this action cost
could be determined by the sentence generator, allowing a deep
co-optimization between instruction planning and sentence generation
(which is future work, see Section~\ref{sec:conclusion}). For now, we
have realized this kind of combination through a simple model
reflecting the fact that referring expressions are easier to generate,
and to understand, for positions adjacent to the last placed block
(which can, for example, be referred to with a pronoun). To this end,
we assume a reduced cost for \taskName{ins-block} if the placed block
is adjacent to the previously placed block (as encoded by the
\factName{lastblock} feature).

%% file: htn-model-complete.tex
\begin{figure}[t]
  \fbox{
\begin{minipage}{\textwidth}
  {\small
    \vspace{-0.4cm}
\begin{align*}
\HObjects = \{&\mathrm{bridge(x, y, z, length, width, orientation), floor(x, y, z, length, width, orientation),} \\&\mathrm{railing(x, y, z, length, orientation), row(x, y, z, length, orientation)} \\&\text{with } \mathrm{x, y, z, length, width \in \mathbb{N}, orientation \in \{north, south, east, west\}}\}\\
\FSetInstruction =~ \{&\textbf{\factName{block(x,y,z)}}, lastblock(x,y,z) \mid x, y, z \in \mathbb{N} \} \cup \{\factName{knows-row}, \factName{knows-floor}, \factName{knows-railing}\} \\
\cTaskNamesInstruction =~ \{&\textbf{\abstractTaskName{build-}}\boldAndIt{X} \mid \boldAndIt{X} \in \HObjects\} \cup  \{\abstractTaskName{ins-build-block}(x,y,z) \mid x,y,z \in \mathbb{N}\} \cup \{\abstractTaskName{ins-build-X} \mid X \in \HObjects\}\\ 
\pTaskNamesInstruction =~ \{&\textbf{\taskName{put-block}(x,y,z)}, \taskName{ins-block}(x,y,z) \mid x,y,z \in \mathbb{N}\} \; \cup \\ \{&\taskName{ins-X}, \taskName{ins-teach-start-X}, \taskName{ins-teach-end-X} \mid X \in \HObjects \setminus \{\mathrm{bridge(x,y,z,l,w,o)}\}  \}\\
\tniInstruction =~ &\abstractTaskName{ins-build-bridge}(0,0,0,5,3,north) 
\end{align*}
\vspace{-1cm}
\begin{align*}
\decmethodsInstruction = \{
\textbf{\abstractTaskName{build-railing}} \boldAndIt{(x,y,z,len,east)} \rightarrow \langle &\taskName{put-block}(x,y,z), \taskName{put-block}(x+len-1,y,z), \\
&\abstractTaskName{build-row}(x+len-1,y,z,len,west) \rangle , \\  
\textbf{\abstractTaskName{build-railing}}\boldAndIt{(x,y,z,len,east)}  \rightarrow \langle &\taskName{put-block}(x,y,z), \abstractTaskName{build-row}(x,y,z,len, east), \\
&\taskName{put-block}(x+len-1,y,z) \rangle , \\
\dots \\
[L]~ \abstractTaskName{ins-build-railing}(x,y,z,len,east) \rightarrow \langle &\abstractTaskName{ins-build-block}(x,y,z), \abstractTaskName{ins-build-block}(x+len-1,y,z), \\&\abstractTaskName{ins-build-row}(x+len-1,y,z,len,west)\rangle,  \\
[H]~ \abstractTaskName{ins-build-railing}(x,y,z,len,east) \rightarrow  \langle&\taskName{ins-railing}(x,y,z,len,east), \abstractTaskName{build-railing}(x,y,z,len,east)\rangle,  \\
[T]~ \abstractTaskName{ins-build-railing}(x,y,z,len,east) \rightarrow \langle&\taskName{ins-teach-start-railing}(x,y,z,len,east), \\&\abstractTaskName{ins-build-block}(x,y,z), \abstractTaskName{ins-build-block}(x+len-1,y,z), \\&\abstractTaskName{ins-build-row}(x+len-1,y,z,len,west), \\&\taskName{ins-teach-end-railing}(x,y,z,len,east)\rangle, \\ 
\dots \\
\abstractTaskName{ins-build-block}(x,y,z) \rightarrow \langle &\taskName{ins-block(x,y,z)}, \taskName{put-block(x,y,z)} \rangle \}
\end{align*}
\noindent \vspace{-0.65cm}
\begin{center}
\begin{tabular} {r|ccc}
  $\pTaskNamesInstruction$
  & \precond & \add & \del \\ \midrule
  \textbf{\taskName{put-block(x,y,z)}} & \{$\neg$ \factName{block(x,y,z)}\}  & \{\factName{block(x,y,z)}\} & \{\}\\
  \taskName{ins-block(x,y,z)} & \{\} & \{\factName{lastblock(x,y,z)}\} &  \{\factName{lastblock(x',y',z')} $\mid \forall x',y',z' \in \mathbb{N}$\}  \\
  \taskName{ins-X}  & \{\factName{knows-T}\} & \{\}  & \{\}\\
  \taskName{ins-teach-start-X}  & \{\}  & \{\}  & \{\} \\
  \taskName{ins-teach-end-X}  & \{\} & \{\factName{knows-T}\} & \{\} \\
 \multicolumn{4}{c}{Here, \factName{knows-T} is the feature where $T$ is $X$'s type, \eg{} $T=$ \factName{railing} for $X=$ railing(x, y, z, length, orientation)}
\end{tabular}
\\
\end{center}
}
\end{minipage}}
\caption{Illustration of our construction-planning model (\textbf{bold
    face}) and instruction-planning model.}
\label{fig:complete_model}
\end{figure}

%% file: instruction-plan-example.tex
\begin{figure}[t]
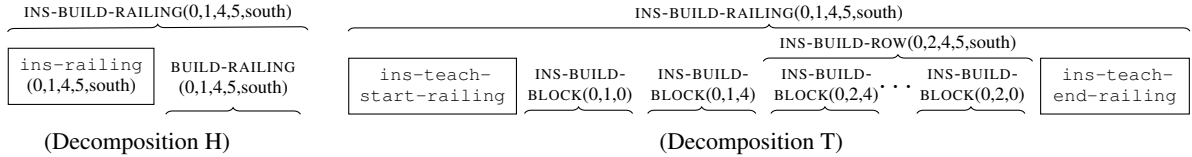

  \captionsetup[subfigure]{labelformat=empty}

  \centering
  \begin{subfigure}{0.25\textwidth}
      \centering
    \includestandalone[]{tikz/railing-high}
    \caption{(Decomposition H)}
  \end{subfigure}
  \quad
    \begin{subfigure}{0.7\textwidth}
      \centering
    \includestandalone[]{tikz/railing-teach}
    \caption{(Decomposition T)}
  \end{subfigure}
  \caption{Two example instruction plans for
    \abstractTaskName{ins-build-railing}(0,1,4,5,south). Abstract tasks at the bottom
    layer need to be further decomposed: \abstractTaskName{Build-railing} could be
    decomposed as in Fig.~\ref{fig:instruction_plan_diagram:low-level}, and
    \abstractTaskName{Ins-build-block} is always decomposed into an \taskName{ins-block}
    and a \taskName{put-block} action.  }
  \label{fig:alternative-decompositions}
\end{figure}

Fig.~\ref{fig:alternative-decompositions} illustrates how the HTN instruction model
interleaves instruction and construction actions to produce valid plans at different
levels of abstraction. If Decomposition H is used, the resulting plan will contain a
single \taskName{ins-railing} instruction action. This will result in sentences such as
(c) or (e) in Fig.~\ref{fig:railing}, depending on context. If Decomposition T is used
instead, the resulting plan will be much more detailed, with one instruction for every
single block that needs to be placed, \eg (b) in Fig.~\ref{fig:railing}.

%% file: tikz/railing-high.tex
\tikzstyle{constructionstep}=[draw,  rectangle, text width=1.15cm, align=center, font=\scriptsize]
\tikzstyle{instructionstep}=[draw, rectangle, text width=1.15cm, align=center, font=\scriptsize]

\newenvironment{nscenter}
 {\parskip=-2pt\par\nopagebreak\centering}
 {\par\noindent\ignorespacesafterend}

\begin{tikzpicture} [node distance = 2cm]
  \node [instructionstep,text width=1.66cm] (C1) {\taskName{ins-railing} (0,1,4,5,south)};
  \node [text width=1.7cm,font=\scriptsize] (R1) [right of = C1] {\begin{nscenter}\abstractTaskName{build-railing} (0,1,4,5,south)\end{nscenter}};

  \draw[decoration={brace},decorate] ($(R1.south west)+(0.1,-0.1)$)  -- ($(R1.south east)+(-0.1,-0.1)$);


  \draw[decoration={brace,raise=6pt},decorate] (C1.north west)  -- node[above=6pt, font=\scriptsize] (R2) {\abstractTaskName{ins-build-railing}(0,1,4,5,south)} (R1.north east);

\end{tikzpicture}

%% file: tikz/railing-teach.tex
\newcommand*\justify{%
  \fontdimen2\font=0.4em
  \fontdimen3\font=0.2em
  \fontdimen4\font=0.1em
  \fontdimen7\font=0.1em
  \hyphenchar\font=`\-
}

\tikzstyle{instructionstep}=[draw, rectangle, align=center, font=\scriptsize]
\tikzstyle{buildblock}=[text width=1.35cm, align=center, font=\scriptsize]

\begin{tikzpicture} 
  \node [instructionstep,text width=1.95cm] (begin)  {\taskName{\justify{ins-teach-start-railing}}};
  \node [buildblock] (C2)  [right = 0cm of begin] {\abstractTaskName{ins-build-block}(0,1,0)};
  \draw[decoration={brace,raise=0pt},decorate] ($(C2.south west)+(0.1, 0)$)  --  ($(C2.south east)+(-0.1, 0)$) ;
  
  \node [buildblock] (C3) [right  = 0cm of  C2] {\abstractTaskName{ins-build-block}(0,1,4)};
  \draw[decoration={brace,raise=0pt},decorate] ($(C3.south west) + (0.1, 0)$) --  ($(C3.south east)+(-0.1, 0)$) ;
  
  \node [buildblock] (R1) [right = 0cm of C3] {\abstractTaskName{ins-build-block}(0,2,4)};
  \draw[decoration={brace,raise=0pt},decorate] ($(R1.south west)+(0.1, 0)$)  --  ($(R1.south east)+(-0.1, 0)$) ;
  
  \node [buildblock] (R2) [right = 0.3cm of R1] {\abstractTaskName{ins-build-block}(0,2,0)};
  \draw[decoration={brace,raise=0pt},decorate] ($(R2.south west)+(0.1,0)$)  --  ($(R2.south east)+(-0.1, 0)$) ;
  \node []  at ($(R1)!0.5!(R2)$) {\bf $\dots$};
  \draw[decoration={brace,raise=-1pt},decorate] (R1.north west) -- node[above=-1pt, font=\scriptsize] (row) {\abstractTaskName{ins-build-row}(0,2,4,5,south)} (R2.north east);

  \node [instructionstep,text width=1.7cm] (end) [right = 0.1cm of R2]   {\texttt{\justify{ins-teach-end-railing}}};
  
  \draw[decoration={brace,raise=10pt},decorate] (begin.north west) -- node[above=10pt, font=\scriptsize] (R1) {\abstractTaskName{ins-build-railing}(0,1,4,5,south)} (end.north east);

\end{tikzpicture}

%% file: experiments.tex
\section{Evaluation}
\label{sec:experiments}

We evaluated the three abstraction-level strategies against each other
on two scenarios.

\subsection{Experimental Setup}
\label{sec:experiment-setup}

\paragraph{Data collection.} We collected evaluation data by asking
human subjects to build complex objects in Minecraft under the
instruction of our IG system. Study participants were recruited
through Prolific. Each participant played a single game, matched to
one of the six conditions (three strategies $\times$ two
scenarios). Participants were required to be fluent in English and own
a Minecraft license. We obtained 20 to 25 plays per condition and paid
each participant $\sim$10 GBP per hour.

In each game, the participant was first informed about the target structure
(``Welcome! I will try to instruct you to
build a [house / bridge]'') and then instructed to build this
structure until it either was complete or a ten-minute time limit was
up and the player had placed at least five correct blocks. We told the
participants explicitly that completing the building was not a
prerequisite for getting paid, to reduce the risk of people quitting
the study because of bad instructions. Either way, the participant was
given a secret code word to enter after the game, to guard against
cheating.

Each participant filled out a post-experiment questionnaire (see
Appendix) after finishing the game. We only
considered games for which we also obtained a questionnaire for the
evaluation.

\paragraph{Scenarios.} We designed two different scenarios to test the
different abstraction strategies. The \emph{house} consists of four
walls, which are each four blocks wide and two block high, and four
rows of four blocks each as the roof.  The house is very minimal and
has neither a door nor windows. We hypothesized that the high-level
strategy would be effective in this scenario, because walls and rows
are commonplace objects which the IF may know without needing to be
taught.

In the \emph{bridge} scenario, subjects were asked to construct the
bridge in Fig.~\ref{fig:railing}, which consists of three complex
objects: the floor and two railings. The railings were specifically
designed to be of a non-obvious shape for participants. Because of this,
we hypothesized that the teaching strategy would work best.

\paragraph{Implementation details.} We used the MC-Saar-Instruct
platform \cite{kohn20:_mc_saar_instr} to connect the IG system to the
study participants. They played Minecraft on their own computers and
connected to our Minecraft server, which then
forwarded their actions to the IG system and the instructions back to
the participants.
Instruction plans were computed offline, to make the different games
comparable and to ensure responsiveness of the IG system (computing a
plan usually takes $<1$ second
but sometimes takes up to $7$ seconds). The natural-language
instructions were computed online, and changed as a function of the state of the
world (\eg{}, referring expressions used different spatial
relations depending on the IF's position in the world).
Whenever the IF places a block incorrectly, \ie{} in a position that
is not consistent with the instruction plan, the IG system needs to guide the IF
back on track.\footnote{While the planner generates a specific reference sequence
  of put-block actions for each object, the IG accepts every possible
  order of put-block actions by the user.} In principle, re-planning methods \cite{GhallabBook}
could be used for this purpose. Here we
opted for a simpler solution: the IG system asks the IF to remove that
block and then returns to the original plan. A similar heuristic
applies when the IF removes a block which was already placed
correctly.

\paragraph{Results.} In addition to the questionnaire, we also
evaluated the IG systems with respect to objective criteria
(percentage of successfully constructed buildings, completion time,
and number of incorrectly placed blocks). The mean results for each
condition are shown in Table \ref{tab:resultsbridge},
including significance test results (Mann–Whitney U test).
Below we also report 95\% bootstrapped confidence intervals (CI).

\input{evaluation-results}

\subsection{Discussion}
\label{sec:bridge-scenario}\label{sec:house-scenario}

\paragraph{Bridge.}
The IFs took significantly longer to build the bridge
under the high-level strategy (mean 275s (CI 217.0, 343.3)) than
under the low-level strategy (177s (CI 154.8, 225.6)), and made
more mistakes (36.9 (CI 24.3, 55.2) vs.\ 18.5 (CI 12.3, 28.1)). This
is because IFs did not know how to build the very specific railing and
thus needed to experiment for a long time. The teaching strategy,
which uses high-level descriptions of complex objects only after first
explaining them, is as fast and accurate as the low-level strategy
(173s (CI 142.7, 226.5)).

At the same time, IFs subjectively rated the low-level strategy
significantly lower on the ``overall'' (2.3 (CI 1.8, 2.7)) and
``clarity'' (2.4 (CI 1.8, 3.0)) questions than the teaching
strategy (overall: 3.4 (CI 2.8, 3.8); clarity: 3.2 (CI 2.7,
3.6)). This confirms our starting hypothesis that low-level
instructions can be perceived as tedious by users. Thus
the teaching system strikes a good balance of task efficiency and
user satisfaction in the bridge scenario.

Looking at the mean building times for the complex subobjects of the
bridge, we found that the high-level strategy is much worse for the
first railing (low-level 49s; teaching 44s; high-level 129s), but
high-level and teaching actually outperform low-level on the second
railing (low-level 38s; teaching 18s; high-level 13s). Thus the use of
high-level explanations can lead to improved task efficiency, if the
complex objects have been explained sufficiently. 

\paragraph{House.} In the ``house'' scenario, the high-level strategy
still has a significantly higher mean task completion time than the
low-level strategy (244s (CI 195.5, 304.2) vs.\ 171s (CI 152.6, 203.1))
and a higher number of mistakes (29.5 (CI 19.6, 51.3) vs.\ 14.5 (CI
10.7, 18.4)). This seems to contradict our original hypothesis that IFs
should be able to process high-level instructions even without
explanation because walls and rows are familiar objects. Furthermore,
unlike in the ``bridge'' scenario, the teaching strategy is slower
than the low-level strategy (239s (CI 195.4, 308.0)) and neither judged
better overall (low: 2.9 (CI 2.4, 3.5), teach: 2.6 (CI 2.1, 3.1)) nor on
clarity (low: 2.8 (CI 2.4, 3.2), teach: 2.4 (CI 2.0, 2.7)). This is
puzzling -- why should teaching the walls slow the IF down in a way
that teaching the railings does not?

To answer this question, we analyzed the building times of the complex
subobjects of the house. We find that the mean completion time for the
four walls is actually \emph{lowest} for the high-level strategy (76s,
compared to low-level 97s, teaching 102s), confirming after all our
hypothesis that high-level instructions are efficient for familiar
complex objects. Where the teaching and high-level strategy fall behind is
the completion time for the first two of the four rows that make up
the roof of the house (low-level 53s, teaching 110s, high-level
155s). A closer inspection of the data suggests that this is because
the instruction the sentence generator computed for the
\taskName{ins-row} action, while semantically and syntactically
correct, is hard to understand (``build a row to the right of length
four to the top of the back right corner of the previous wall''). The
low-level block-by-block strategy does not have that problem. Thus,
the sentence generator for high abstraction levels must be
tested and designed with special care.

%% file: evaluation-results.tex
\newcommand{\nonsig}{\textsuperscript{\fullmoon}}
\newcommand{\marsig}{\textsuperscript{{\color{gray}\newmoon}}}
\newcommand{\strsig}{\textsuperscript{\newmoon}}

\begin{table}
  \centering   \small
\begin{tabular}[]{lrrr @{\hspace{2em}}rrr}
  \toprule
&  \multicolumn{3}{c}{Bridge} &   \multicolumn{3}{c}{House}\\
  \cmidrule(lr{2em}){2-4}  \cmidrule(l{-0.5em}r){5-7}
& Low-level & Teaching & High-level & Low-level & Teaching & High-level\\
  \midrule
  Success rate (\%)                   & 95 & 100 & 90         & 100 & 95 & 89      \\
  Time to success (seconds)            & 177.0 & 172.5\nonsig & 275.5\strsig\strsig & 171.6 & 239.8\marsig & 244\marsig\nonsig\\
  Number of mistakes made              & 18.5 & 18.6\nonsig & 36.9\marsig\strsig & 14.5 & 23.3\nonsig & 29.5\marsig\nonsig\\\midrule
  I had to re-read instructions.       & 4.7 & 4.1\marsig & 4.6\nonsig\marsig & 3.9 & 4.4\nonsig & 4.6\marsig\nonsig\\
  It was always clear what to do.      & 2.4 & 3.2\strsig & 1.9\nonsig\strsig & 2.8 & 2.4\nonsig & 1.9\strsig\nonsig\\
  Overall gave good instructions.      & 2.3 & 3.4\strsig & 2.4\nonsig\strsig & 2.9 & 2.6\nonsig & 2\strsig\marsig\\
  Gave useful feedback about progress.\hspace{-2em} & 3.9 & 3.8\nonsig & 4.1\nonsig\nonsig & 4.1 & 3.9\nonsig & 4\nonsig\nonsig\\
  System was really verbose.           & 2.2 & 2.8\marsig & 2.2\nonsig\nonsig & 2.2 & 2.9\strsig & 2.9\nonsig\nonsig\\
  Instructions were too early.         & 3.8 & 3.5\nonsig & 3.1\marsig\nonsig & 3.2 & 3\nonsig & 3.8\nonsig\marsig\\
  Instructions were too late.          & 1.4 & 1.7\nonsig & 2.1\strsig\nonsig & 1.4 & 2\strsig & 1.6\nonsig\marsig\\
  \bottomrule
\end{tabular}

\caption{Means of evaluation results; all data points except ``success rate'' are only
  from successful games.  Questions results are on a 1-5 scale (1; disagree completely, 5:
  agree completely).  Statements slightly shortened, full text in appendix.  Significance
  levels (Mann-Whitney U) comparing to low-level (for teaching) / low-level and teaching
  (for high-level) shown as \strsig{}: \(p<0.05\), \marsig{}: \(p<0.1\), \nonsig{}:
  \(p>0.1\).  }
\label{tab:resultsbridge}
\end{table}

%% file: conclusion.tex
\section{Conclusion}
\label{sec:conclusion}

\vspace{-0.05cm}

We have shown how to generate building instructions in Minecraft at
different levels of abstraction through HTN planning. Our evaluation
shows that the level of abstraction matters to human users, and that
our ``teaching'' strategy strikes an effective balance of high-level
and low-level instructions.

We view our system as the first step towards a more general
framework. We used a very basic model of the user's knowledge; in
future work, we will train statistical models on user interactions to
estimate what a user knows initially and what they learn during the
experiment. Furthermore, we used hand-designed action costs. Perhaps
the most interesting avenue of future research is to let the sentence
generator specify the action costs, based on how easy it is to express
a given instruction action in language in the current world and
dialogue state. This will allow the sentence generator to influence
the search of the discourse planner, offering a new method for closing
the ``generation gap'' \cite{10.5555/100711}.

%% file: appendix.tex
\section{The Post Game Questionnaire}
\label{sec:post-game-quest}

\begin{itemize}
\item Overall, the system gave me good instructions.
\item I had to re-read instructions to understand what I needed to do.
\item It was always clear to me what I was supposed to do.
\item The system’s instructions came too late or too early.
\item The system was really verbose and explained things that were already clear to me.
\item The system gave me useful feedback about my progress.
\item Please add any comments or observations you had (free text)
\end{itemize}
All but the last question are five-point Likert scale questions (Disagree completely -- Agree completely).

\section{Build Times for the Bridge Scenario}
\label{sec:build-times-bridge}

\begin{tabular}{lr@{ }lr@{ }lr@{ }l}
\toprule
& \multicolumn{2}{c}{Low-level} & \multicolumn{2}{c}{Teaching} & \multicolumn{2}{c}{High-level} \\
\midrule
floor & 84.1 &(CI 71.6 100.5) & 103.9 &(CI 78.5 164.5) & 128.9 &(CI 91.6 190.5)\\
railing 1 & 49.4 &(CI 40.3 68.2) & 44.3 &(CI 37 53.1) & 129.9 &(CI 92.4 190.4)\\
railing 2 & 38.2 &(CI 24.3 79.1) & 18.4 &(CI 10.7 47.4) & 13.3 &(CI 9.6 21.6)\\
\bottomrule
\end{tabular}

\section{Build Times for the House Scenario}
\label{sec:build-times-house}

\begin{tabular}{lr@{ }lr@{ }lr@{ }l}
  \toprule
  & \multicolumn{2}{c}{Low-level} & \multicolumn{2}{c}{Teaching} & \multicolumn{2}{c}{High-level} \\
\midrule
wall 1 & 51.9 &(CI 46.3 61.2) & 68.6 &(CI 59.1 78.2) & 52.9 &(CI 39.9 68.1)\\
wall 2 & 23.2 &(CI 18 34.4) & 15.2 &(CI 11.7 24) & 10.3 &(CI 8.1 16.3)\\
wall 3 & 12.7 &(CI 10.5 16.1) & 8.6 &(CI 6.5 11.9) & 6.7 &(CI 5.4 10.1)\\
wall 4 & 9.4 &(CI 8.1 10.8) & 9.5 &(CI 6 17.4) & 5.8 &(CI 4.5 7.6)\\
row 1 & 17.1 &(CI 14 22.7) & 42.7 &(CI 32.3 68.4) & 69.6 &(CI 52.5 95.1)\\
row 2 & 35.5 &(CI 21.5 61.5) & 66.9 &(CI 38.7 124.6) & 85.2 &(CI 48.7 153.9)\\
row 3 & 9.0 &(CI 6.2 14.5) & 8.6 &(CI 5.8 17.6) & 7.9 &(CI 4.1 18.1)\\
row 4 & 5.7 &(CI 4.3 9.9) & 13.9 &(CI 4.9 44.6) & 3.8 &(CI 2.8 5.2)\\
\bottomrule
\end{tabular}

\section{Planning Times for the different Configurations}
\label{sec:planning-times}

\begin{tabular}{lrrr}
	\toprule
	Scenario & Low-level & Teaching & High-level\\
	\midrule
	Bridge & 0.78s & 0.81s & 0.82s\\
	House & 5.92s & 0.96s & 6.98s\\
	\bottomrule
\end{tabular}